\documentclass{article}
\usepackage{spconf,amsmath,graphicx,amsfonts,xcolor,url,booktabs,multirow,comment, makecell, float}

\title{Combining Acoustics, Content and Interaction Features to Find \mbox{Hot Spots} in Meetings}
\name{Dave Makhervaks${}^{1\ast}$ \quad 
   William Hinthorn${}^2$ \quad
   Dimitrios Dimitriadis${}^2$ \quad
   Andreas Stolcke${}^{3\ast}$ %
  \thanks{${}^\ast$Research done while the authors were with Microsoft}}
 
 \address{${}^1$Dept.~of Computer Science, Harvey Mudd College, Claremont, CA, USA \\
	${}^2$Speech and Dialog Research Group, Microsoft, Redmond, WA, USA \\
	${}^3$Amazon Alexa Speech, Sunnyvale, CA, USA \\
	{\tt\small dmakhervaks@g.hmc.edu \quad \{wihintho,didimit\}@microsoft.com \quad stolcke@amazon.com} }

\begin{document}
\ninept
\sloppy

\maketitle
\begin{abstract}
Involvement hot spots have been proposed as a useful concept for meeting analysis and studied off and on for over 15 years. 
These are regions of meetings that are marked by high participant involvement, as judged by human annotators.
However, prior work was either not conducted in a formal machine learning setting, or focused on only a subset of possible meeting features or downstream applications (such as summarization).
In this paper we investigate to what extent various acoustic, linguistic and pragmatic aspects of the meetings, both in isolation and jointly, can help detect hot spots. In this context, the openSMILE toolkit \cite{opensmile} is to used to extract features based on acoustic-prosodic cues, BERT word embeddings \cite{BERT} are used for encoding the lexical content, and a variety of statistics based on speech activity are used to describe the verbal interaction among participants.
In experiments on the annotated ICSI meeting corpus, we find that the lexical model is the most informative, with incremental contributions from interaction and acoustic-prosodic model components.
\end{abstract}

\section{Introduction and Prior Work}
\label{sec:intro}

A definition of meeting ``hot spots'' was first introduced in \cite{WredeShriberg:2003}, where it was investigated whether human annotators could reliably identify regions in which participants are ``highly involved in the discussion.''
The motivation was that meetings generally have low information density and are tedious to review verbatim after the fact. 
An automatic system that could detect regions of high interest (as indicated by the involvement of the participants {\em during} the meeting) would thus be useful.
Relatedly, automatic meeting summarization could also benefit from such information by giving extra weight to hot spot regions in selecting or abstracting material for inclusion in the summary.
Later work on the relationship between involvement and summarization \cite{LaiEtAl:interspeech2013} defined a different approach: hot spots are those regions chosen for inclusion in a summary by human annotators (``summarization hot spots'').

In the present work we stick with the original ``involvement hot spot'' notion, and refer to such regions simply as ``hot spots'', regardless of their possible role in summarization.
We note that high involvement may be triggered both by a meeting's content (``what is being talked about'' or ``what may be included in a textual summary''), as well as behavioral and social factors, such as a desire to participate, to stake out a position, or to oppose another participant.
A related notion in dialog system research is ``level of interest'' \cite{SchullerEtAl:icslp2006}.

The initial research on hot spots focused on the reliability of human annotators and correlations with certain low-level acoustic features, such as pitch \cite{WredeShriberg:2003}. Also investigated were the correlation between hot spots and dialog acts \cite{WredeShriberg:asru2003} and hot spots and speaker overlap \cite{CetinShriberg:mlmi2006}, without however conducting experiments in automatic hot spot prediction using machine learning techniques.
Laskowski \cite{Laskowski:slt2008} redefined the hot spot annotations in terms of time-based windows over meetings, and investigated various classifier models to detect ``hotness'' (i.e., elevated involvement).
However, that work focused on only two types of speech features: presence of laughter and the temporal patterns of speech activity across the various participants, both of which were found to be predictive of involvement.

For the related problem of level-of-interest prediction in dialog systems \cite{WangHirschberg:sigdial2011}, it was found that content-based classification can also be effective, using both a discriminative TF-IDF model and lexical affect scores, as well as prosodic features.
In line with the earlier hot spot research on interaction patterns and speaker overlap,
 turn-taking features were shown to be helpful for spotting summarization hot spots, and even more so than the human involvement annotations \cite{LaiEtAl:interspeech2013}.
The latter result confirms our intuition that summarization-worthiness and involvement are different notions of ``hotness.''


In this paper, following Laskowski, we focus on the automatic prediction of the speakers' involvement in sliding-time windows/segments.
We evaluate machine learning models based on a range of features that can be extracted automatically from audio recordings, either directly via signal processing or via the use of automatic transcriptions (ASR outputs).
In particular, we investigate the relative contributions of three classes of information: 
\begin{itemize}
    \item low-level acoustic-prosodic features, such as those commonly used in other paralinguistic tasks, like sentiment analysis (extracted using openSMILE \cite{opensmile});
    \item spoken word content, as encoded with a state-of-the-art lexical embedding approach such as BERT \cite{BERT};
    \item speaker interaction, based on speech activity over time and across different speakers.
\end{itemize}
We attach lower importance to laughter, even though it was found to be highly predictive of involvement in the ICSI corpus, partly because we believe it would not transfer well to more general types of (e.g., business) meetings, and partly because laughter detection is still a hard problem in itself \cite{Gosztolya:2018}.
Generation of speaker-attributed meeting transcriptions, on the other hand, has seen remarkable progress \cite{Denmark:interspeech2019} and could support the features we focus on here.

\section{Data}
\label{sec:data}

\label{sec:overview}
The ICSI Meeting Corpus \cite{Janin03} is a collection of meeting recordings that has been thoroughly annotated, including annotations for involvement hot spots \cite{HSguide:2005}, linguistic utterance units, and word time boundaries based on forced alignment.
The dataset is comprised of 75 meetings with a real-time audio duration of about 70 hours.
Meetings have six speakers on average. 
Most of the participants are well-acquainted and friendly with each other.
Hot spots were originally annotated with 8 levels and degrees, ranging from `not hot' to `lukewarm' to `hot +'. Every utterance was labeled with one of these discrete labels by a single annotator.
Hightened involvement is rare, being marked on only 1\% of utterances.

Due to the severe imbalance in the label distribution, Laskowski \cite{Jin_+04} proposed extending the involvement, or hotness, labels to sliding time windows.
In our implementation (details below), this resulted in 21.7\% of samples (windows) being labeled as ``involved''.

We split the corpus into three subsets: training, development, and evaluation, keeping meetings intact.  Table \ref{tab:table1} gives statistics of these partitions.


\begin{table}[htb]
\caption{Partitions of the ICSI dataset}
\begin{minipage}[b]{1.0\linewidth}
  \begin{center}
    \label{tab:table1}
    \begin{tabular}{|l|c|c|c|}
    \hline
      \textbf{Purpose} & {Training} & {Development} & {Evaluation}\\
      \hline
      \textbf{\# Meetings} & 51 & 9 & 15\\
      \hline
      \textbf{\# Words} & 478,593 & 120,533 & 147,478\\
      \hline
      \textbf{\# Utterances} & 69,755 & 18,360 & 21,283\\
      \hline
      \textbf{\# Windows} & 10,197 & 2,452 & 3,174\\
      \hline
      \textbf{Share of hot windows} & 21.5\% & 22.3\% & 21.9\% \\ \hline
    \end{tabular}
  \end{center}
  \end{minipage}
\end{table}

We were concerned with the relatively small number of meetings in the test sets, 
and repeated several of our experiments with a (jackknifing) cross-validation setup over the training set.  The results obtained were very similar to those with the
fixed train/test split results that we report here.

\subsection{Time Windowing}
\label{sec:slidingwindows}
\begin{figure}[tb]
\begin{minipage}[b]{1.0\linewidth}
  \centering
  \centerline{\includegraphics[width=8.5cm]{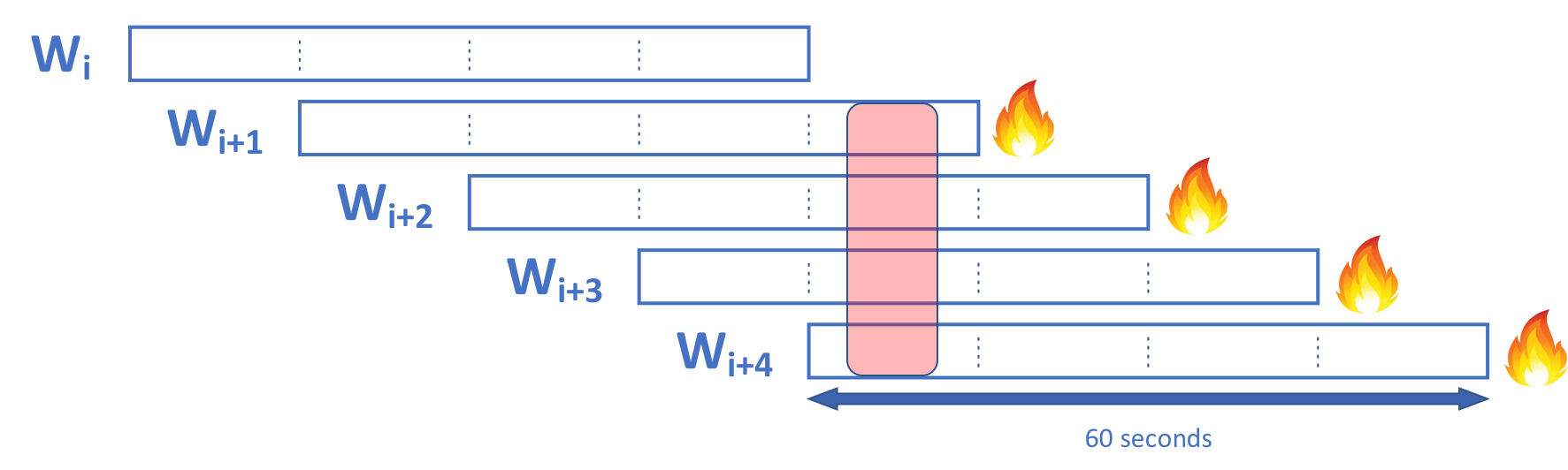}}
\end{minipage}
\caption{Visualization of the sliding window defining data points. The area bounded by the red box indicates labeled involvement, causing 4 windows to be marked as `hot'.}
\label{fig:res}
\end{figure}

As stated above, the corpus was originally labeled for hot spots at the utterance level, where involvement was marked by either a `b' or a `b+' label.
Training and test samples for our experiments correspond to
 60\,s-long sliding windows, with a 15\,s step size.
 If a certain window, e.g., a segment spanning the times 15\,s \ldots 75\,s, overlaps with any involved speech utterance, then we  label that whole window as `hot'.
 Fig.~\ref{fig:res} gives a visual representation. 
 
\subsection{Metric}

In spite of the windowing approach, the class distribution is still skewed, and an accuracy metric would reflect the particular class distribution in our data set.
Therefore, we adopt the unweighted average recall (UAR) metric commonly used in emotion classification research.
UAR is a reweighted accuracy where the samples of both classes are weighted equally in aggregate.  UAR thus simulates a uniform class distribution.
To match the objective, our classifiers are trained on appropriately weighted training data.
Note that chance performance for UAR is by definition 50\%, making results more comparable across different data sets and tasks.

\section{Feature Description}
\label{sec:pagestyle}

\subsection{Acoustic-Prosodic Features}
Prosody encompasses pitch, energy, and durational features of speech. Prosody is thought to convey emphasis, sentiment, and emotion, all of which are presumably correlated with expressions of involvement. 
We used the openSMILE toolkit \cite{opensmile} to compute 988 features as defined by the {\it emobase988} configuration file, operating on the close-talking meeting recordings.
This feature set consists of low-level descriptors such as intensity, loudness, Mel-frequency cepstral coefficients, and pitch. For each low-level descriptor, functionals such as max/min value, mean, standard deviation, kurtosis, and skewness are computed. Finally, global mean and variance normalization are applied to each feature, using training set statistics.
The feature vector thus captures acoustic-prosodic features aggregated over what are typically utterances.
We tried extracting openSMILE features directly from 60\,s windows, but found better 
results by extracting subwindows of 5\,s, followed by pooling over the longer 60\,s duration.
We attribute this to the fact that emobase features are designed to operate on individual utterances, which have durations closer to 5\,s than 60\,s.

\subsection{Word-Based Features} 
\subsubsection{Bag of words with TF-IDF}
Initially, we investigated a simple bag-of-words model including  all unigrams, bigrams, and trigrams found in the training set. Occurrences of the top 10,000 n-grams were encoded to form a 10,000-dimensional vector, with values weighted according to TD-IDF.
TF-IDF weights n-grams according to both their frequency (TF) and their salience (inverse document frequency, IDF) in the data, where each utterance was treated as a separate document.
The resulting feature vectors are very sparse.
 
\subsubsection{Embeddings}
The ICSI dataset is too small to train a neural embedding model from scratch.
Therefore, it is convenient to use the pre-trained BERT embedding architecture \cite{BERT} to create an utterance-level embedding vector for each region of interest.
Having been trained on a large text corpus, the resulting embeddings encode semantic similarities among utterances, and would enable generalization from word patterns seen in the ICSI training data to those that have not been observed on that limited corpus.

We had previously also created an adapted version of the BERT model,  tuned to perform utterance-level sentiment classification, on a separate dataset~\cite{BDS_19}.
As proposed in~\cite{BERT}, we fine-tuned all layers of the pre-trained BERT model by adding a single fully-connected layer and classifying using only the embedding corresponding to the classification ([CLS]) token prepended to each utterance. The difference in UAR between the hot spot classifiers using the pre-trained embeddings and those using the sentiment-adapted embeddings is small. Since the classifier using embeddings extracted by the sentiment-adapted model yielded slightly better performance, we report all results using these as input.

To obtain a single embedding for each 60 s window, we experimented with various approaches of pooling the token and utterance-level embeddings. For our first approach, we ignored the ground-truth utterance segmentation and speaker information. We merged all words spoken within a particular window into a single contiguous span. Following~\cite{BERT}, we added the appropriate classification and separation tokens to the text and selected the embedding corresponding to the [CLS] token as the window-level embedding.  
Our second approach used the ground-truth segmentation of the dialogue. Each speaker turn was independently modeled, and utterance-level embeddings were extracted using the representation corresponding to the [CLS] token. Utterances that cross window boundaries are truncated using the word timestamps, so only words spoken within the given time window are considered.  For all reported experiments, we use L2-norm pooling to form the window-level embeddings for the final classifier, as this performed better than either mean or max pooling.


\subsection{Speaker Activity Features}
These features were a compilation of three different feature types:

\emph{Speaker overlap percentages}: Based on the available word-level times, we computed a 6-dimensional feature vector, where the $i$th component indicates the fraction of time that $i$ or more speakers are talking within a given window. This can be expressed by $\frac{t_i}{60}$ with $t_i$ indicating the time in seconds that $i$ or more people were speaking at the same time.

\emph{Unique speaker count}: Counts the unique speakers within a window, as a useful metric to track the diversity of participation within a certain window.

\emph{Turn switch count}: Counts the number of times a speaker begins talking within a window. This is a metric similar to the number of utterances. However, unlike utterance count, turn switches can be computed entirely from speech activity, without requiring a linguistic segmentation.

\subsection{Laughter Count}
Laskowski found that laughter is highly predictive of involvement in the ICSI data \cite{Laskowski:slt2008}.
Laughter is annotated on an utterance level and falls into two categories: laughter solely on its own (no words) or laughter contained within an utterance (i.e., while speaking). The feature is a simple tally of the number of times people laughed within a window. 
We include it in some of our experiments for comparison purposes, though we do not trust it as general feature. (The participants in the ICSI meetings are far too familiar and at ease with each other to be representative with regards to laughter.)

\section{Modeling}


\subsection{Non-Neural Models}
In preliminary experiments, we compared several non-neural classifiers, including 
logistic regression (LR), random forests, linear support vector machines, and multinomial naive Bayes.
Logistic regression gave the best results all around, and we used it exclusively for the results shown here, unless neural networks are used instead.

\subsection{Feed-Forward Neural Networks}
\label{subsection:NN}
\subsubsection{Pooling Techniques}
For BERT and openSMILE vector classification, we designed two different feed-forward neural network architectures. The  sentiment-adapted embeddings
described in Section~\ref{sec:pagestyle} produce one 1024-dimensional vector per utterance.
Since all classification operates on time windows, we had to pool over all utterances falling withing a given window, taking care to truncate words falling outside the window.
We tested four pooling methods: L2-norm, mean, max, and min, with L2-norm giving the best results.

As for the prosodic model, each vector extracted from openSMILE represents a 5\,s interval. Since there was both a channel/speaker-axis and a time-axis, we needed to pool over both dimensions in order to have a single vector representing the prosodic features of a 60\,s window.  The second to last layer is the pooling layer,  max-pooling across all the channels, and then mean-pooling over time. The output of the pooling layer is directly fed into the classifier.

\subsubsection{Hyperparameters}
The hyperparameters of the neural networks (hidden layer number and sizes) were also tuned in preliminary experiments.
Details are given in Section~\ref{sec:experiments}.

\subsection{Model Fusion}

\begin{figure}[tb]
\begin{minipage}[b]{1.0\linewidth}
  \centering
  \centerline{\includegraphics[width=8.5cm]{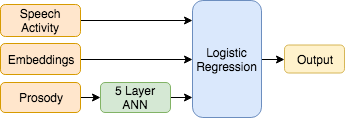}}
\end{minipage}
\caption{Overview of fusion model}
    \label{fig:fusion}
\end{figure}

Fig.~\ref{fig:fusion} depicts the way features from multiple categories are combined.
Speech activity and word features are fed directly into a final LR step.
Acoustic-prosodic features are first combined in a feed-forward neural classifier, whose 
output log posteriors are in turn fed into the LR step for fusion.
(When using only prosodic features, the ANN outputs are used directly.)

\section{Experiments}
\label{sec:experiments}
We group experiments by the type of feaures they are based on:
 acoustic-prosodic, word-based, and speech activity, evaluating each group first by itself, and then in combination with others.  

\subsection{Speech Feature Results}
As discussed in Section \ref{sec:pagestyle}, a multitude of input features were investigated,  with some being more discriminative. The most useful speech activity features were speaker overlap percentage, number of unique speakers, and number of turn switches, giving evaluation set UARs of 63.5\%, 63.9\%, and 66.6\%, respectively.
When combined the UAR improved to 68.0\%, showing that these features are partly complementary.

\subsection{Word-Based Results}
The TF-IDF model alone gave a UAR of 59.8\%. A drastic increase in performance to 70.5\% was found when using the BERT embeddings instead.  
Therefore we adopted embeddings for all further experiments based on word information.

Three different types of embeddings were investigated, i.e. sentiment-adapted embeddings at an utterance-level, unadapted embeddings at the utterance-level, and unadapted embeddings over time windows. 

The adapted embeddings (on utterances) performed best, indicating that adaptation to sentiment task is useful for involvement classification. It is important to note, however, that the utterance-level embeddings are larger than the window-level embeddings. This is due to there being more utterances than windows in the meeting corpus.

The best neural architecture we found for these embeddings is a 5-layer neural network with sizes 1024-64-32-12-2. Other  hyperparameters for this model are dropout rate = 0.4, learning rate = $10^{-7}$ and tanh activation function.
The UAR on the evaluation set with just BERT embeddings as input was 65.2\%.

Interestingly, the neural model was outperformed by a LR directly on the component values of the embedding vectors, with a UAR of 70.5\%.
Perhaps the neural network requires further fine-tuning, or the neural model is too prone to overfitting, given the small training corpus. 
In any case, we use LR on embeddings for all subsequent results.

\subsection{Acoustic-Prosodic Feature Results}
Our prosodic model is a 5-layer ANN,  as described in Section~\ref{subsection:NN}. The architecture is: 988-512-128-16-Pool-2. The hyperparameters are: dropout rate $= 0.4$, learning rate $= 10^{-7}$, tanh activation. The UAR on the evaluation set with just openSMILE features was 62.0\%. 

\subsection{Fusion Results and Discussion}

\begin{table}[tb]
    \centering
    \caption{Hot spot classification results with individual feature subsets, all features, and with individual feature sets left out.}
    \label{tab:results-all}
    \begin{tabular}{l|c|c}
        \textbf{Feature Set} &   \textbf{UAR w/ Features}   &   \textbf{UAR w/o Features} \\
        \hline
        Prosody     &   62.0\%            & 71.7\%      \\
        Speech-act  &   68.0\%            & 72.2\%     \\
        Words       &   70.5\%             & 68.4\%      \\
        \hline
        All        &   72.6\%            &   N/A     \\
    \end{tabular}
\end{table}

Table~\ref{tab:results-all} gives the UAR for each feature subset individually, for all features combined, and for a combination in which one feature subset in turn is left out.
The one-feature-set-at-a-time results suggest that prosody, speech activity and words are of increasing importance in that order.
The leave-one-out analysis agrees that the words are the most important (largest drop in accuracy when removed), but on that criterion the prosodic features are more important than speech-activity.
The combination of all features was 0.4\% absolute better than any other subset, showing that all feature subsets are non-redundant.

\begin{figure}[tb]
\begin{minipage}[b]{1.0\linewidth}
  \centering
  \centerline{\includegraphics[width=8.5cm]{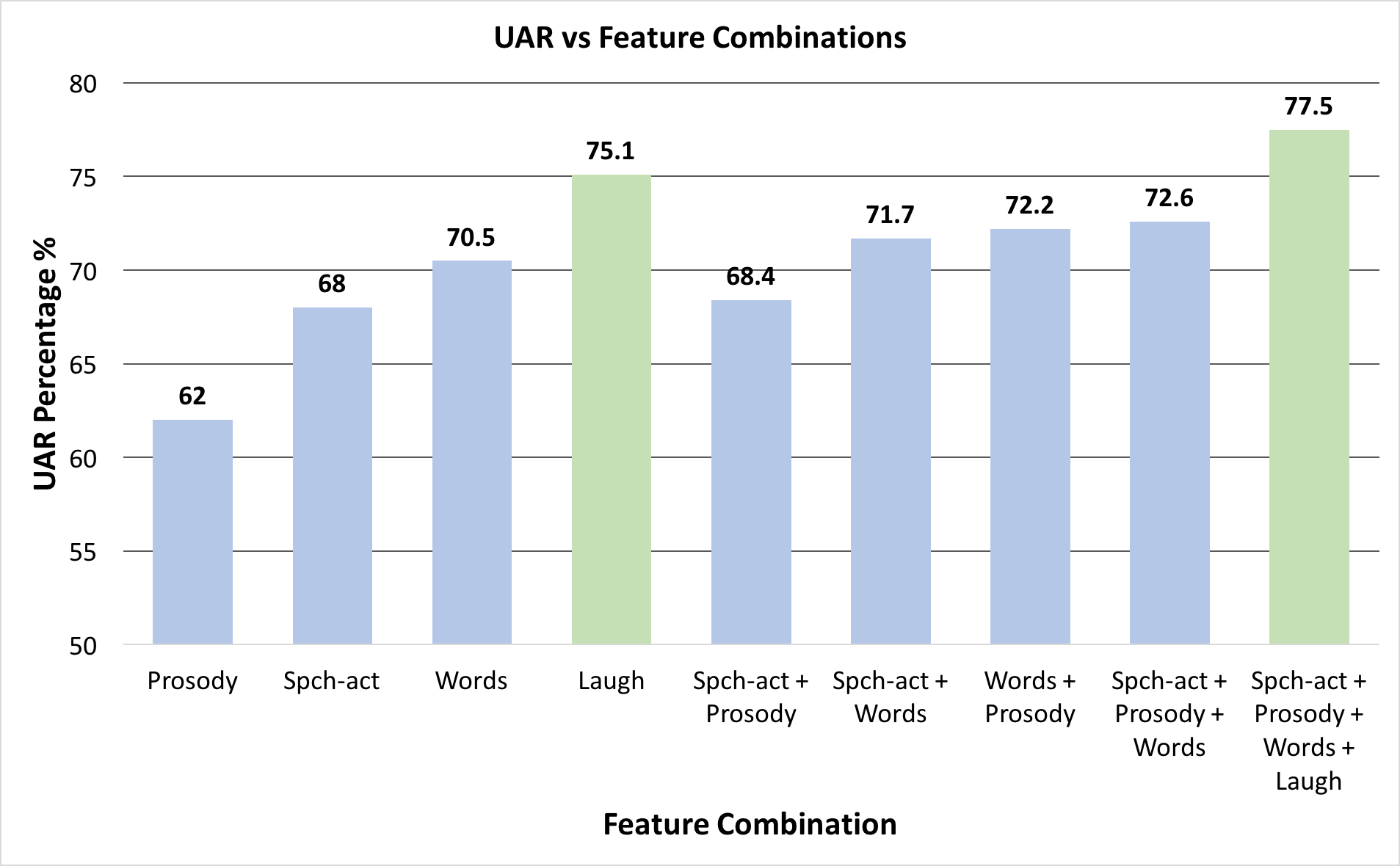}}
\end{minipage}
\caption{Graph of different combinations of features. Green rectangles indicate models using laughter. Prosody = openSMILE features with NN, Words = embeddings, Spch-act = speech activity, Laugh = laughter count. Combination was by logistic regression.}
    \label{fig:results-all}
\end{figure}



Fig.~\ref{fig:results-all} shows the same results in histogram form, 
but also includes those with laughter information.
Laughter count by itself is the strongest cue to involvement, as Laskowski \cite{Laskowski:slt2008} had found.
However, even given the strong individual laughter feature, the other features add information, pushing the UAR from from 75.1\% to 77.5\%.

\section{Conclusion}
We studied detection of areas of high involvement, or ``hot spots'', within meetings using the ICSI corpus. The features that yielded the best results are in line with our intuitions.  Word embeddings, speech activity features such as number of turn changes, and prosodic features are all plausible and effective indicators of high involvement. 
Furthermore, the feature sets are partly complementary and yield best results
when combined using a simple logistic regression model.
The combined model achieves 72.6\% UAR, or 77.5\% with laughter feature.

For future work, we would want to see a validation on an independent meeting collection, such as business meetings.
Some features, in particular laughter, may not be as useful in this case.
More data could also enable the training of joint models that perform an early fusion of the different feature types.
Also, the present study still relied on human transcripts, and it would be important to know how much UAR suffers with a realistic amount of speech recognition error.
Transcription errors are expected to boost the importance of the features types that do not rely on words.



\section{Acknowledgments}
We thank Britta Wrede, Elizabeth Shriberg and Kornel Laskowski for
sharing details about the data and annotations.



\ninept
\bibliographystyle{IEEEbib}
\bibliography{refs}

\end{document}